\documentclass[10pt, conference, letterpaper]{IEEEtran}
\IEEEoverridecommandlockouts
\usepackage{cite}
\usepackage{amsmath,amssymb,amsfonts}
\usepackage{algorithmic}
\usepackage{graphicx}
\usepackage{textcomp}
\usepackage{xcolor}
\usepackage{tabularx}
\usepackage{hyperref}
\usepackage{url}
\usepackage[labelformat=simple]{subcaption}
\def\BibTeX{{\rm B\kern-.05em{\sc i\kern-.025em b}\kern-.08em
    T\kern-.1667em\lower.7ex\hbox{E}\kern-.125emX}}
\begin{document}

\title{
MADQRL: Distributed Quantum Reinforcement Learning Framework for Multi-Agent Environments
} 


\author{\centering
\IEEEauthorblockN{Abhishek Sawaika\textsuperscript{1}, Samuel Yen-Chi~Chen\textsuperscript{2},
Udaya Parampalli\textsuperscript{1}, Rajkumar Buyya\textsuperscript{1}}\\
\IEEEauthorblockA{
\textsuperscript{1}{Quantum Cloud Computing and Distributed Systems Lab (qCLOUDS), School of Computing and Information Systems},\\
{The University of Melbourne, Australia}\\
\textsuperscript{2}{Brookhaven National Laboratory, Upton, NY, USA}\\
\textit{asawaika@student.unimelb.edu.au; ycchen1989@ieee.org; (udaya, rbuyya)@unimelb.edu.au}
}
}

\maketitle

\begin{abstract}
Reinforcement learning (RL) is one of the most practical ways to learn from real-life use-cases. Motivated from the cognitive methods used by humans makes it a widely acceptable strategy in the field of artificial intelligence. Most of the environments used for RL are often high-dimensional, and traditional RL algorithms becomes computationally expensive and challenging to effectively learn from such systems. Recent advancements in practical demonstration of quantum computing (QC) theories, such as compact encoding, enhanced representation and learning algorithms, random sampling, or the inherent stochastic nature of quantum systems, have opened up new directions to tackle these challenges. Quantum reinforcement learning (QRL) is seeking significant traction over the past few years. However, the current state of quantum hardware is not enough to cater for such high-dimensional environments with complex multi-agent setup. To tackle this issue, we propose a distributed framework for QRL where multiple agents learn independently, distributing the load of joint training from individual machines. Our method works well for environments with disjoint sets of action and observation spaces, but can also be extended to other systems with reasonable approximations.  We analyze the proposed method on cooperative-pong environment and our results indicate $\sim$10\% improvement from other distribution strategies, and $\sim5$\% improvement from classical models of policy representation.

\end{abstract}

\section{Introduction}

Quantum computing has attracted considerable attention due to its ability to provide computational advantages over classical computing for certain classes of problems \cite{nielsen2010quantum}. Although present-day quantum hardware is constrained by noise, limited coherence times, and device imperfections, extensive research efforts are underway to realize practical quantum advantages \cite{Corcoles2020ChallengesSystems}. Within this context, quantum machine learning (QML) has emerged as a particularly promising direction, aiming to integrate quantum computational principles into machine learning (ML) frameworks. Most existing QML approaches adopt a hybrid quantum–classical architecture \cite{McCaskey2018HybridSystems}, where quantum processors are used for subroutines that can exploit quantum effects, while classical processors manage components better suited to conventional computation.
Variational quantum algorithms (VQAs) \cite{Cerezo2021VariationalAlgorithms} constitute the core computational paradigm underlying most contemporary QML models. Built upon parameterized quantum circuits optimized via classical routines, VQAs have enabled QML methods to be applied across a broad range of learning scenarios, including supervised classification \cite{Schuld2018SupervisedComputers}, sequential and time-series modeling \cite{chen2022quantumLSTM, Sawaika2025ADetection}, natural language processing \cite{li2023pqlm,yang2022bert,di2022dawn,stein2023applying}, and graph based learning \cite{Innan2024FinancialNetworks}, etc.

Within this landscape, QRL \cite{Chen2026QuantumApplications} has recently emerged as a particularly compelling research direction, aiming to exploit quantum-enhanced representations, optimization dynamics, and stochasticity to improve the decision-making capabilities of learning agents. As real-world environments increasingly involve decentralized decision processes, partial observability, and interactions among multiple autonomous entities, extending QRL frameworks beyond single-agent settings toward multi-agent quantum reinforcement learning (MA-QRL) becomes a natural and necessary progression \cite{Yu2023QuantumDirections}. Such multi-agent formulations open new opportunities to investigate quantum-enabled coordination, competition, and information sharing among agents, while also posing fundamental challenges related to scalability.

%
To enable such systems and overcome the scalability issues on near-term quantum devices, we propose a distributed framework for such MA-QRL problems. Major contributions of our work include:
\begin{enumerate}
    \item An architecture for distributed quantum reinforcement learning for multi-agent environments, learning joint tasks through independent training. 
    \item Detailed analysis of this framework to showcase practical usability and limitations on a benchmarking environment of two agent cooperative pong game.
\end{enumerate}


The rest of the paper is organised as follows: A brief introduction to RL in Section \ref{sec: RL}, followed by our solution methodology in Section \ref{sec: SM}, and discussions on experiments and results in Section \ref{sec: exp and results}. Finally, we conclude and provide remarks on possible future directions in Section \ref{sec; conclude}.

\section{Reinforcement Learning}
\label{sec: RL}

Reinforcement learning is a class of ML problems that aims to learn an optimal strategy for state-action pairs (policy), from an environment usually described through a Markov decision process (MDP) \cite{Puterman1990ChapterProcesses}. Training usually involves an iterative feedback-based loop, where an agent (learner) continuously interacts with the environment (MDP) to gradually learn the policy, while maximizing the expected reward using a controlled exploration-exploitation mechanism \cite{Sutton1999BookLearning}.


\subsection{Single-Agent Environments}
Consider a simplistic representation of a RL problem using the following MDP definition:

\begin{enumerate}
    \item \textbf{State Space (S) }: $ S = S_O \cup S_I \cup S_T $, is a collection of starting ($S_O$), intermediate ($S_I$) and terminating ($S_T$) states.
    \item \textbf{Action Space (A) }: $A = \{a_1, a_2, ...\}$, is the set of possible actions.
    \item \textbf{Transition Function (P)} : $S^1 \times A \times S^2 \rightarrow [0,1] $, defines the probability of transition $P(s,a,s') = Pr(s' | s,a)$ from a state '$s$' to a state '$s'$' under an action '$a$'.
    \item \textbf{Reward Function (R)} : $S^1 \times A \times S^2 \rightarrow \mathbb{R} $, defines the reward for a transition $(s,a,s')$ from state '$s$' to a state '$s'$' under an action '$a$'.
\end{enumerate}

Where, $(S^1, \; S^2 )\subseteq S, \; s \in S^1, \; s' \in S^2 \; \text{and} \; a \in A$. Then a single-agent environment (SAE) is the one where the entire action space (A) is managed by only one learning agent. 


\subsection{Multi-Agent Environments}
\label{MAE}

A multi-agent environment (MAE) is an extension of the previous definition, where the entire action space (A) can be divided into disjoint sets of action spaces ($A_i$), s.t.

$A_i = \{a_{i1}, a_{i2}, ...\} \; | \; A = \bigcup A_i \; \text{and} \;  A_i \cap A_j = \phi, \; \forall i \neq j$. Where, each action space $A_i$ is associated with a distinct learning agent (i). Correspondingly, a transition $(s,\textbf{a},s')$ is defined by a joint action vector $\textbf{a} : \times A_i = (a_{1x}, a_{2y}, ... , a_{iz}), \; \forall i>1 \; \text{and} \; \exists a_{i\alpha} \in A_i$.

Some of the popular MAE include games like Chess, GO, Pong, etc., and real world applications like robotics, power systems, autonomous vehicles, etc.\cite{Busoniu2008ALearning}. These can be modeled either in a collaborative system (COL-MAE) where agents train towards a common objective, or a competitive system (COMP-MAE) with conflicting rewards \cite{Kraus1997NegotiationEnvironments}.

\subsection{Deep Reinforcement Learning}

Given the environment description, the objective of a RL algorithm is to learn a policy, s.t.:

\textbf{Policy} $\boldsymbol{(\pi)} : S \times A \rightarrow [0,1] $ defines the posterior probability of selecting an action '$a$' given the current state '$s$' i.e. $\pi(s,a) = Pr(a|s)$,

while maximizing the \textbf{expected discounted reward}:

\begin{equation}
    \underset{\pi}{max}\Bigg(\mathbb{E}_{s_t \sim P, a_t \sim \pi}\bigg[\sum_{t=0}^{t=\infty}{\gamma^t*R(s_t,a_t,s_{t+1})}\bigg]\Bigg)
    \label{eq: objective}
\end{equation}

Where, $0 < \gamma < 1$ is the discount factor to gradually decrease the impact of rewards coming from far-off future. 

Most of the practical environments are often high-dimensional, and it becomes challenging to solve for such objective using traditional RL algorithms such as temporal-difference (TD) \cite{tesauro1995temporal} or Q-Learning \cite{Watkins1992Q-learning}. Therefore, the learnable functions like Policy ($\pi$), Value ($V_\pi(s)$) and Quality ($Q_\pi(s,a)$) can be represented as deep neural networks and some suitable algorithm can be used to learn these mappings. This is the central idea behind deep reinforcement learning (DRL) \cite{Li2017DeepOverview}.

Here, \textbf{Value Function} $\boldsymbol{(V_\pi)} : S \rightarrow \mathbb{R}$, defines the expected reward of starting from state '$s$' and following the policy '$\pi$' from time $t > 0$, s.t. $V_\pi(s) = $
$$\mathbb{E}_{s_t \sim P, a_t \sim \pi}\bigg[\sum_{t=0}^{t=\infty}{\gamma^t*R(s_t,a_t,s_{t+1}) \mid s_o = s, \; \pi}\bigg]$$

and \textbf{Quality Function} $\boldsymbol{(Q_\pi)} : S \times A \rightarrow \mathbb{R}$, provides the expected reward of starting from state '$s$' with action '$a$' and then following the policy '$\pi$' from time $t > 0$, s.t. $Q_\pi(s,a) = $
$$\mathbb{E}_{s_t \sim P, a_t \sim \pi}\bigg[\sum_{t=0}^{t=\infty}{\gamma^t*R(s_t,a_t,s_{t+1}) \mid s_o = s, \; a_o=a, \; \pi}\bigg]$$

\subsection{PPO Learning Algorithm}

Proximal Policy Optimization (PPO) algorithm \cite{Schulman2017ProximalAlgorithms} is a special case of policy gradient method \cite{Sutton1999PolicyApproximation}, where a parametrized policy ($\boldsymbol{\pi(\theta)}$) is learned to find the optimal values of parameters $\boldsymbol{(\theta)}$ for an objective $\boldsymbol{J(\theta)}$, similar to that in Equation \ref{eq: objective}, over a finite horizon ($0 \leq t < T$), using gradient ascent techniques. i.e. $\theta_{i+1} = \theta_i + \alpha\nabla_\theta * J(\theta)$.

The objective function specific to PPO algorithm is defined as:
\begin{equation}
    J(\theta)  = \underset{\theta}{max}\big(\mathbb{E}_{t} \left[
        \min \left(
            SO, \; SO\_clipped
        \right)
    \right]\big)
    \label{eq:ppo objective}
\end{equation}

Where, $SO$ = $r_{t}(\theta) * \hat{A}_{t}$, $SO\_clipped$ = $\operatorname{clip}\big(r_{t}(\theta), 1 - \epsilon, 1 + \epsilon\big)*\hat{A}_{t}$, and 

$\hat{A}_{t}$ is the estimate of \textbf{Advantage function} $\boldsymbol{(A_\pi)}$ which calculates the benefit of taking an action '$a$' from a starting state '$s$', s.t.
$$A_\pi(s,a) = Q_\pi(s,a) - V_\pi(s)$$
The policy ratio $r_t(\theta) = \pi_{\theta}(a_t | s_t) \; / \; \pi_{\theta_{\text{old}}}(a_t | s_t)$, identifies the importance of action '$a$' between the current and the previous policy, and hence guides the direction of gradient movement in the learning process. Authors in PPO had introduced this clipping technique, see Equation \ref{eq:ppo objective}, to restrict $r_t(\theta)$ within an $\epsilon$ bound and thereby avoiding unbalanced updates of policy parameters. 

\section{Solution Methodology}
\label{sec: SM}

This section presents our  proposed distributed learning framework designed for MA-QRL problems, along with the details on model architecture used for policy representation. This system is designed specifically for large COL-MAE system and works well for environments with distinct observation spaces and less dependency on information from other agents. 

\subsection{Distributed Learning Framework}

\begin{figure*}[h]
    \vskip -0.1in
    \centering
\includegraphics[width=.9\linewidth]{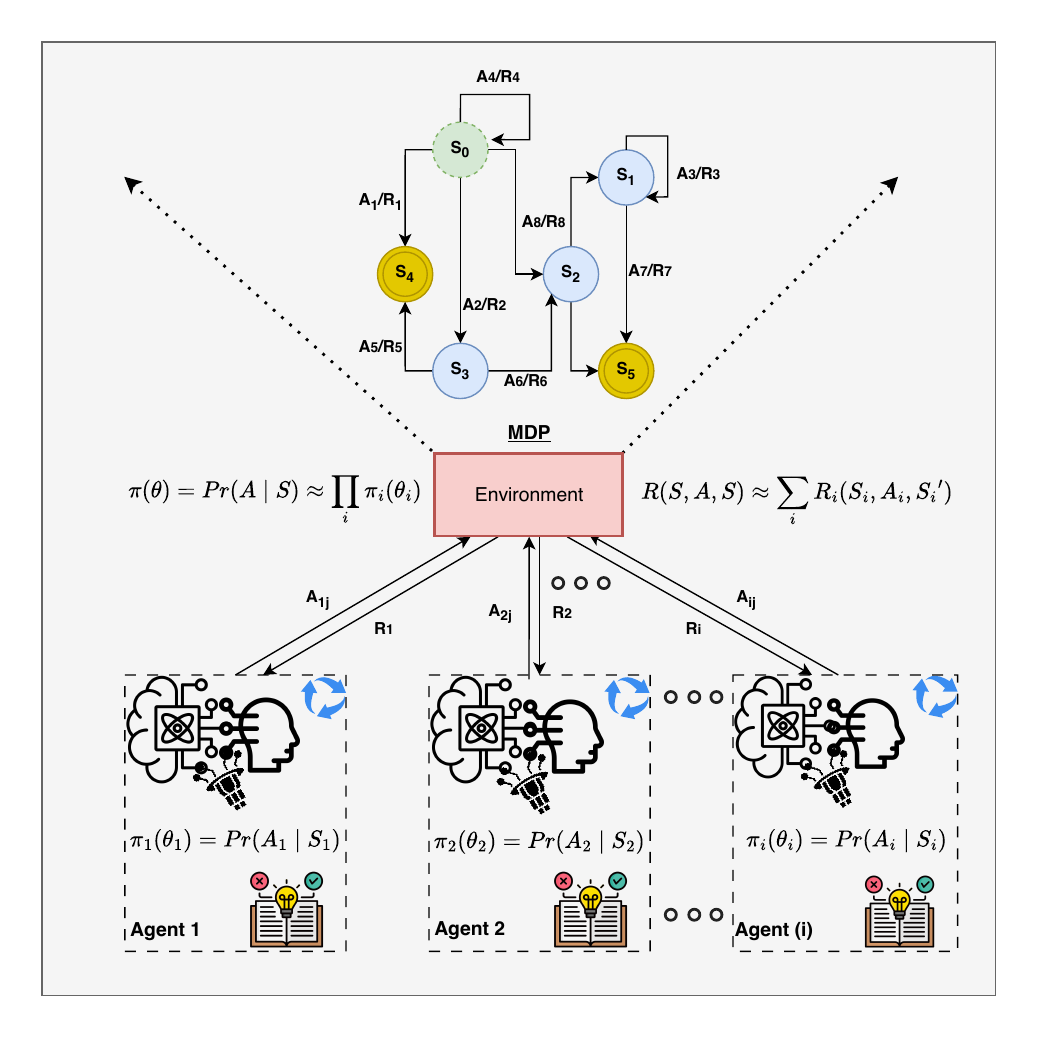}
    \vskip -0.15in
    \caption{ {\bfseries Architecture of MADQRL framework.} Agents learn independently to optimize for joint objective defined by the environment. This independent training happens on local observation, reward and action spaces, such that the joint policy can be approximated by the product of local policies. Here, the MDP representation of the environment is just an example for completeness of the system.}
    \label{fig:arch}
    \vskip -0.1in
\end{figure*}

Authors in \cite{Gronauer2021Multi-agentSurvey} have used three classes to classify learning strategies for MAE, namely centralized training centralized execution (CTCE), distributed training decentralized execution (DTDE) and centralized training and decentralized execution (CTDE). In this work we have used the following nomenclatures, with CTCE as "Joint", DTDE as "Independent" and CTDE as "Shared" for our experiments.

CTCE trains a joint policy $$\pi^{CTCE}(\theta) = Pr(\boldsymbol{a} \mid \boldsymbol{s})$$
where \textbf{a} is the joint action space of all agents performing simultaneous transitions $(\boldsymbol{s}, \boldsymbol{a} , \boldsymbol{s'})_\pi$ from joint states $\boldsymbol{s}$ to $\boldsymbol{s'}$, as described in Section \ref{MAE}. 

Whereas, in CTDE every agent learns a shared policy 
$$\pi^{CTDE}(\theta) \simeq Pr(\boldsymbol{a} \mid s_i) = \prod\pi_i(\theta)*I(i,j), \; \forall j\neq i$$
which is trained over local states '$s_i$' and mutual information 'I'. $\pi_i(\theta)$ is similar to that in $\pi_i^{DTDE}(\theta)$.

At last, through DTDE every agent (i) learns a local policy $$\pi_i^{DTDE}(\theta) = Pr(a_i \mid s_i)$$ 
trained over local states '$s_i$' and actions '$a_i$, as if they are learning independently. The Joint policy $\boldsymbol{\pi(\theta)}$ can be written by
$$Pr(\boldsymbol{a} \mid \boldsymbol{s}) \approx \prod_{i} \limits Pr(a_i \mid s_i)$$
We took motivation from these strategies and designed a framework for distributed quantum reinforcement learning with hybrid quantum-classical model, within the DTDE class. It is opposed to the recent work in \cite{Chen2025Quantum-Train-BasedLearning} which train multiple copies of a SAE and periodically share policy parameters for weight synchronization. This simply distributes the training load to multiple learners rather than learning a multi-agent environment using distributed training.

Figure \ref{fig:arch} presents the architecture diagram of our proposed framework for distributed quantum reinforcement learning for multi-agent environments (MADQRL). 

\subsection{Quantum Policy Representation}
\label{QRL}

Recent research in QRL has shown promise in the use of various quantum-inspired techniques for different RL tasks \cite{Chen2026QuantumApplications}. Where, authors in \cite{Dong2008QuantumLearning} encoded the MDP into quantum states and proposed a projective measurement \cite{Barberena2024OverviewMeasurements} and Grover's search \cite{Grover1996ASearch} based strategy for a TD like policy update, authors in \cite{Chen2020VariationalLearning} uses VQC and QNN based architectures for policy representation in a DRL based setting.

A variational quantum circuit (VQC) usually consists of three major components:
\begin{enumerate}
    \item An\textbf{ Encoder Circuit $\boldsymbol{E(X)}$}: to encode classical data into quantum bits.
    \item A\textbf{ Parametrized Circuit $\boldsymbol{C(\Theta_q)}$}: for learning user defined objectives.
    \item \textbf{Measurements and Post Processing}: to translate quantum information into classical output / objectives for optimization.
\end{enumerate}


We design a quantum neural network (QNN) \cite{Gupta2001QuantumNetworks} based model, with three layers of hybrid quantum-classical layer, to represent the policy function $\pi(\theta)$. Here, the policy parameters $\theta$ are directly mapped to QNN's parameters $\Theta = \Theta_q \cup \theta_c$, where $\Theta_q \; \text{and} \; \theta_c $ are quantum and classical model parameters, respectively. Each agent has a separate version of the policy model for training. i.e. $\pi_i(\theta_i) \mid \theta_i \rightarrow \Theta_i$.

A single layer of the QNN model consists of the following layers:

\begin{table}[h]
    \vskip -0.15in
    \centering
    \begin{tabular}{|c|}\hline
       Flatten \\\hline
       Linear\_Layer \\\hline 
       VQC \\\hline 
       RELU \\\hline
       Linear\_layer \\\hline
        RELU\\ \hline
    \end{tabular}
    \label{tab:Layers}
    \vskip -0.05in
\end{table}

\begin{figure}[h]
    \centering
    \includegraphics[width=\linewidth, height = 200pt]{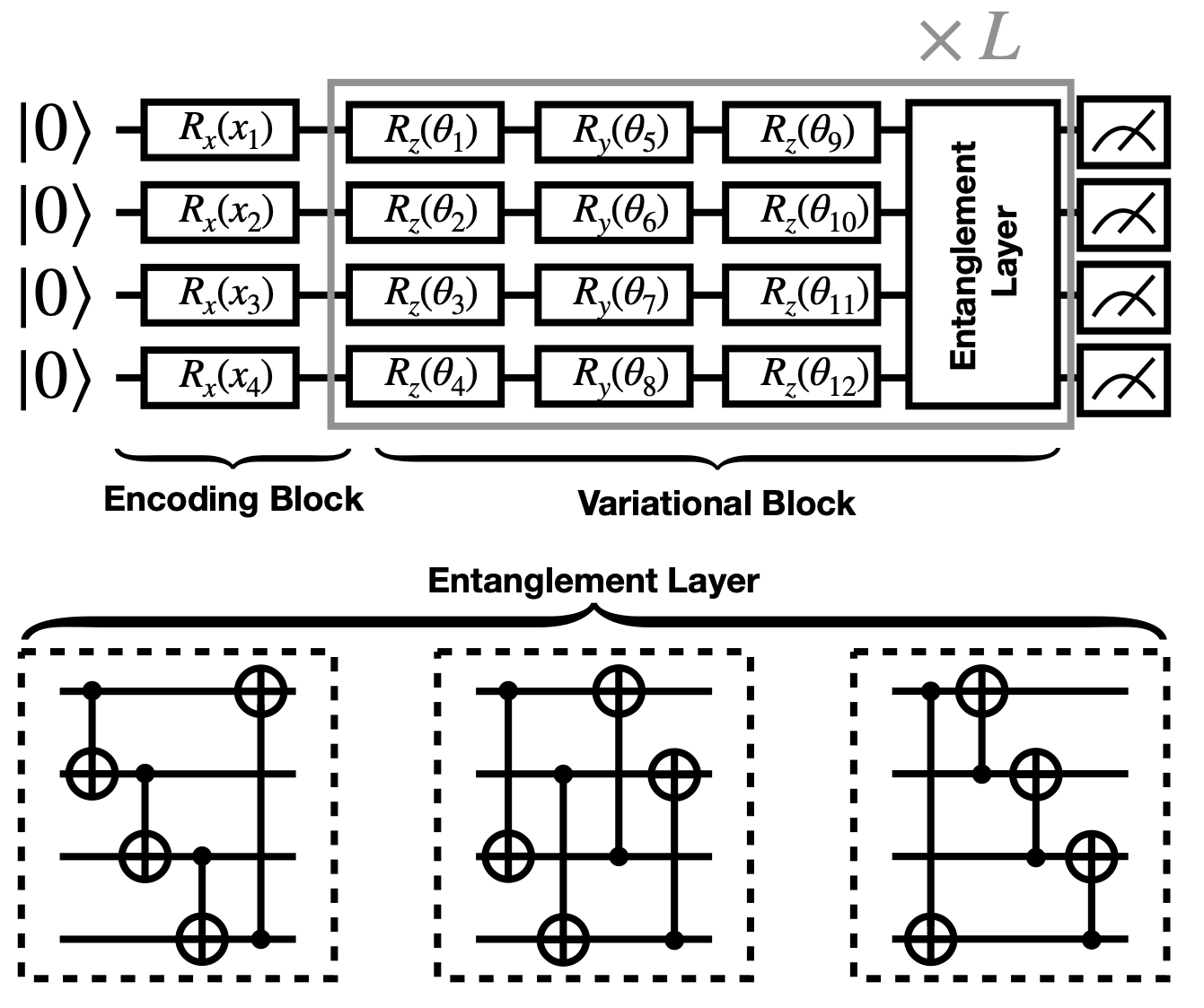}
    \caption{A sample quantum circuit with 4 qubits, having angle encoding using RX gates and $L = 3$ layers of strongly entangled variational layers, followed by full measurement in $Z$-Basis. Here we use three different entanglement layers.}
    \label{fig:program}
    \vskip -0.15in
\end{figure}

We use angle encoding with Rx gate and multiple layers of fully-entangled parametrized circuit with sequence of Rx, Rz and CNOT gates in the VQC model. This design help us to access larger sections of the Hilbert space and provide maximal entanglement properties for efficient policy training. It also helps in increasing the expressivity power of learning process with smaller model sizes. Figure \ref{fig:program} shows a sample VQC with 4 qubit and 3 layers of parametrized circuit.

\section{Performance Evaluation}
\label{sec: exp and results}
We present the experimental setup used for our study and the analysis of our findings. The designed system can work with any traditional learning algorithm for a DRL based system, but for our experiments we have used PPO for policy update.

\subsection{Experiment Design}
\begin{figure}[h]
    \centering
    \includegraphics[width=0.6\linewidth]{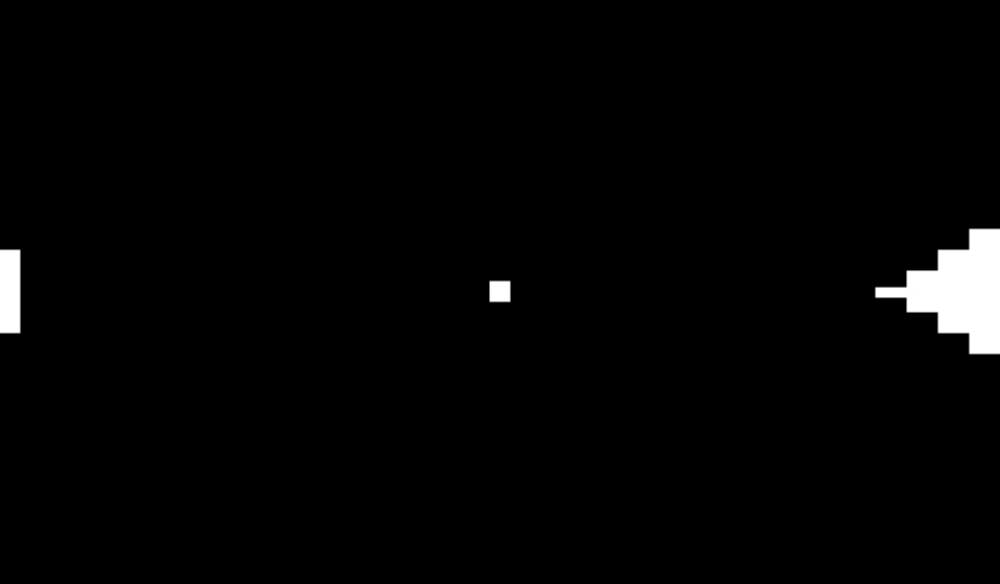}
    \caption{A snapshot of the cooperative-pong game environment. The entire black box is the "playing arena", the small white dot at the center is the "ball", which moves freely based on the laws of physics, and the two elongated white area on the left and the right edges are the paddles, representing agents. Here, the right one is a cake shaped paddle, while the one on the left side is a standard block paddle.}
    \label{fig:env}
    \vskip -0.05in
\end{figure}

We experiment with cooperative-pong \cite{pong} environment provided by PettingZoo \cite{terry2021pettingzoo}. This environment has 2 agents with discrete action space: \{-1,0,1\}, representing \{left, no, right\} movements, respectively. The aim of this game is to keep the ball within the playing arena for longest period of time, by precise movements of paddles (agents), based on ball dynamics. Fig \ref{fig:env} shows a sample snapshot of the game. It is important to note that the ball can leave the arena only from the left or the right edges. The state space is represented by an rgb array of size (560 x 960 x 3), and each agent observe only half of the screen. We scaled the canvas size (state space) to 64 x 64 for simplicity of our experiments. 

We choose this game because it is a COL-MAE and the observation space per agent are disjoint. 
Therefore, the joint policy $\pi(\theta)$ can be factorized as $\pi(\theta) = \pi_1(\theta_1)*\pi_2(\theta_2)$, s.t. optimizing for the individual rewards jointly optimize for the whole game, i.e. the ball remain in the court for longer durations.

We run the whole training pipeline using Ray framework \cite{222605_ray} for $\sim$15K iterations, with batch size of 512, learning rate ($\alpha$) = $10^{-4}$, $\epsilon$ = 0.3, $\gamma = 0.95$, kl coefficient = 0.2, vf coefficient = 1.0 and entropy coefficient = 0.5. We compare performance between the quantum model, as described in Section \ref{QRL}, with 13 qubits and 9 layers, and the classical model with CNN \cite{Chauhan2018ConvolutionalRecognition} layers activated by RELU functions.

\subsection{Results Analysis}

We conduct experiments to compare the proposed framework of "independent" training against two other strategies of "joint" and "shared" learning. We test for both classical and quantum models. We also compare the effect of entanglement on training of the quantum model in joint learning setup, within a basic entangling layer (BE) and a strongly entangling layer (SE).

\begin{figure}[h]
    \centering
    \includegraphics[width=.9\linewidth]{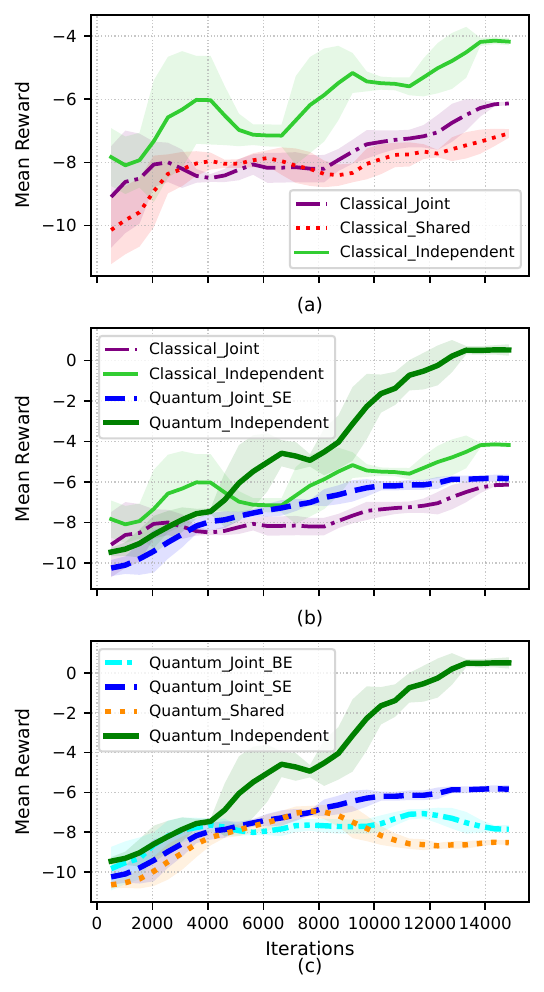}
    \caption{Comparing learning curves for different distributed learning frameworks between (a) classical, (b) quantum–classical, and (c) quantum models. Results are reported for a 2-agent cooperative Pong game with the observation space reduced to $64 \times 64$. Curves show the moving average of the mean reward with a window size of $5$, while the shaded regions indicate the corresponding standard deviation.}
    \label{fig:results_plot}
    \vskip -0.1in
\end{figure}

As mentioned earlier, one of the key contributors to the success of our proposed design is the disjoint nature of observation space, even though the game is collaborative. It is observed from Figure \ref{fig:results_plot} that the independent learning strategy performs best and the shared mechanism performs worst. This is due to unnecessary confusions created for additional information learning and sharing in the shared case. We can also verify this in Table \ref{tab:performance} from the large amount of trainable parameters used in the shared case, and the degradation in test episodes post learning. 

The independent learning strategy takes more time 
exploring, and hence takes more time to train. Joint policy on the other hand takes least time to train upto a reasonable performance. This is measured using saturation point, see Table \ref{tab:performance}, observed while learning with incremental number of trials in the stochastic process of episode sampling.


\begin{table}[h]
    \centering
    \caption{Performance analysis on two agent cooperative pong game for different combinations of policy model and distribution strategies. A CNN and a QNN based architectures are used for classical and quantum models, respectively. There are 1170 tunable parameters in the VQCs used in quantum models.}
    \label{tab:performance}
    \begin{tabular}{|>{\centering\arraybackslash}p{0.18\linewidth}|>{\centering\arraybackslash}p{0.12\linewidth}|>{\centering\arraybackslash}p{0.11\linewidth}|>
    {\centering\arraybackslash}p{0.14\linewidth}|>{\centering\arraybackslash}p{0.19\linewidth}|}
    \hline
        \textbf{Model - Distribution} & \textbf{Total Runtime (s)} & \textbf{Test Episodes} & \textbf{Trainable Weights} & \textbf{Sampling Saturation (\%iterations)}\\\hline
        Classical-Shared & 377 &  193 & 40074 & 40\\\hline
        Classical-Joint & \textbf{291} & 254 & 20560 & 35.8\\\hline
        Classical-Independent & 391 & \textbf{264} & \textbf{18752} & 45\\\hline
        \hline\hline
        Quantum-Shared & 2753 & 176 & 20140 & 37.9 \\\hline
        Quantum-Joint SE & \textbf{1529}& 259 & 10889 & 37.8 \\\hline
        Quantum-Joint BE & 1795 & 245 & 10893 & 37.5 \\\hline
        Quantum-Independent &  2835 & \textbf{288} & \textbf{10691} & 48\\
    \hline
    \end{tabular}
\end{table}

It is important to note that the number of trainable parameters for quantum models are almost half of that of their classical counterparts, yet they perform better in most of the cases, see Table \ref{tab:performance}. Except for the shared case, where the classical model outperforms quantum.

On the contrary, since VQC's are trained on a simulator, there is a trade-off for runtime, where quantum model takes more time due to exponential cost of qubit simulations.

From results in Figure \ref{fig:results_plot}(c), we see that the model with SE layer performs better than the one with BE layer. Where, it has higher average reward during training process, lower runtime and larger test episodes. This is mainly because the quantum circuit with higher entanglement compensates for the information loss due to significant feature reduction by the linear layer used for compressing data for circuits with fewer qubits.

\section{Conclusions and Future Work}
\label{sec; conclude}

We proposed a framework for distributed quantum reinforcement learning using independent training and hybrid quantum-classical model for policy representation. We demonstrated the utility of such design for  environments with disjoint spaces and minimal communication requirements. 

Our experiments on cooperative-pong environment using PPO algorithm showed $\sim$10\% improvement in final test episodes and mean episodic rewards for a hybrid quantum-classical representation. Since VQC's are trained on a simulator, there is a trade-off for training runtime. Moreover, given the nature of the underlying environment, the proposed distributed framework with independent learning outperformed other strategies of joint and shared learning, for both classical and quantum representations.

This work can be easily integrated with other learning algorithms and extended for environments having similar structural but different environment parameters such as, but not limited to, continuous state, action and observation spaces, and for competitive environments where information about other agents are often not known. It can also be applied for environments modeled using partially observable MDPs. Being distributed, it can scale well for larger environments. 

\section*{Acknowledgments}
This work is supported by the University of Melbourne and Maitri scholarships from the Department of Foreign Affairs and Trade, Government of Australia.

\bibliographystyle{ieeetr}
\bibliography{bibliography,references,qml_bib}

@mual{pong,
    title = {{Cooperative Pong - PettingZoo Documentation}, {https://pettingzoo.farama.org/environments/butterfly/cooperative\_pong/}}
}

@article{terry2021pettingzoo,
  title={Pettingzoo: Gym for multi-agent reinforcement learning},
  author={Terry, Jordan and Black, Benjamin and Grammel, Nathaniel and Jayakumar, Mario and Hari, Ananth and Sullivan, Ryan and Santos, Luis S and Dieffendahl, Clemens and Horsch, Caroline and Perez-Vicente, Rodrigo and others},
  journal={Advances in Neural Information Processing Systems},
  volume={34},
  pages={15032--15043},
  year={2021}
}

@article{tesauro1995temporal,
  title={Temporal difference learning and TD-Gammon},
  author={Tesauro, Gerald and others},
  journal={Communications of the ACM},
  volume={38},
  number={3},
  pages={58--68},
  year={1995}
}

@inproceedings {222605_ray,
author = {Philipp Moritz and Robert Nishihara and Stephanie Wang and Alexey Tumanov and Richard Liaw and Eric Liang and Melih Elibol and Zongheng Yang and William Paul and Michael I. Jordan and Ion Stoica},
title = {Ray: A Distributed Framework for Emerging {AI} Applications},
booktitle = {13th USENIX Symposium on Operating Systems Design and Implementation (OSDI 18)},
year = {2018},
isbn = {978-1-939133-08-3},
address = {Carlsbad, CA},
pages = {561--577},
url = {https://www.usenix.org/conference/osdi18/presentation/moritz},
publisher = {USENIX Association},
month = oct
}

@book{nielsen2010quantum,
  title={Quantum computation and quantum information},
  author={Nielsen, Michael A and Chuang, Isaac L},
  year={2010},
  publisher={Cambridge University Press}
}

@inproceedings{chen2022quantumLSTM,
  title={Quantum long short-term memory},
  author={Chen, Samuel Yen-Chi and Yoo, Shinjae and Fang, Yao-Lung L},
  booktitle={2022 IEEE International Conference on Acoustics, Speech and Signal Processing (ICASSP)},
  pages={8622--8626},
  year={2022},
  organization={IEEE}
}

@inproceedings{li2023pqlm,
  title={PQLM-Multilingual Decentralized Portable Quantum Language Model},
  author={Li, Shuyue Stella and Zhang, Xiangyu and Zhou, Shu and Shu, Hongchao and Liang, Ruixing and Liu, Hexin and Garcia, Leibny Paola},
  booktitle={2023 IEEE International Conference on Acoustics, Speech and Signal Processing (ICASSP)},
  pages={1--5},
  year={2023},
  organization={IEEE}
}

@inproceedings{yang2022bert,
  title={When BERT Meets Quantum Temporal Convolution Learning for Text Classification in Heterogeneous Computing},
  author={Yang, Chao-Han Huck and Qi, Jun and Chen, Samuel Yen-Chi and Tsao, Yu and Chen, Pin-Yu},
  booktitle={2022 IEEE International Conference on Acoustics, Speech and Signal Processing (ICASSP)},
  pages={8602--8606},
  year={2022},
  organization={IEEE}
}

@inproceedings{di2022dawn,
  title={The dawn of quantum natural language processing},
  author={Di Sipio, Riccardo and Huang, Jia-Hong and Chen, Samuel Yen-Chi and Mangini, Stefano and Worring, Marcel},
  booktitle={2022 IEEE International Conference on Acoustics, Speech and Signal Processing (ICASSP)},
  pages={8612--8616},
  year={2022},
  organization={IEEE}
}

@inproceedings{stein2023applying,
  title={Applying QNLP to sentiment analysis in finance},
  author={Stein, Jonas and Christ, Ivo and Kraus, Nicolas and Mansky, Maximilian Balthasar and M{\"u}ller, Robert and Linnhoff-Popien, Claudia},
  booktitle={2023 IEEE International Conference on Quantum Computing and Engineering (QCE)},
  volume={2},
  pages={20--25},
  year={2023},
  organization={IEEE}
}

@article{Busoniu2008ALearning,
    title = {{A comprehensive survey of multiagent reinforcement learning}},
    year = {2008},
    journal = {IEEE Transactions on Systems, Man and Cybernetics Part C: Applications and Reviews},
    author = {Bu{\c{s}}oniu, Lucian and Babu{\v{s}}ka, Robert and De Schutter, Bart},
    number = {2},
    month = {3},
    pages = {156--172},
    volume = {38},
    url = {https://ieeexplore.ieee.org/abstract/document/4445757},
    doi = {10.1109/TSMCC.2007.913919},
    issn = {10946977}
}

@article{Grover1996ASearch,
    title = {{A fast quantum mechanical algorithm for database search Citation in BibTeX format A fast quantum mechanical algorithm for database search}},
    year = {1996},
    author = {Grover, Lov K},
    isbn = {0897917855},
    doi = {10.1145/237814.237866}
}

@article{Sawaika2025ADetection,
    title = {{A Privacy-Preserving Federated Framework with Hybrid Quantum-Enhanced Learning for Financial Fraud Detection}},
    year = {2025},
    author = {Sawaika, Abhishek and Krishna, Swetang and Tomar, Tushar and Suggisetti, Durga Pritam and Lal, Aditi and Shrivastav, Tanmaya and Innan, Nouhaila and Shafique, Muhammad},
    month = {7},
    url = {https://arxiv.org/pdf/2507.22908},
    arxivId = {2507.22908}
}

@article{Sutton1999BookLearning,
    title = {{Book Reviews Reinforcement Learning}},
    year = {1999},
    author = {Sutton, R S and Barto, A G},
    isbn = {0262193981}
}

@article{Corcoles2020ChallengesSystems,
    title = {{Challenges and Opportunities of Near-Term Quantum Computing Systems}},
    year = {2020},
    journal = {Proceedings of the IEEE},
    author = {Corcoles, Antonio D. and Kandala, Abhinav and Javadi-Abhari, Ali and McClure, Douglas T. and Cross, Andrew W. and Temme, Kristan and Nation, Paul D. and Steffen, Matthias and Gambetta, Jay M.},
    number = {8},
    month = {8},
    pages = {1338--1352},
    volume = {108},
    publisher = {Institute of Electrical and Electronics Engineers Inc.},
    url = {https://ieeexplore.ieee.org/abstract/document/8936946},
    doi = {10.1109/JPROC.2019.2954005},
    issn = {15582256},
    arxivId = {1910.02894}
}

@article{Puterman1990ChapterProcesses,
    title = {{Chapter 8 Markov decision processes}},
    year = {1990},
    journal = {Handbooks in Operations Research and Management Science},
    author = {Puterman, Martin L.},
    number = {C},
    month = {1},
    pages = {331--434},
    volume = {2},
    publisher = {Elsevier},
    url = {https://www.sciencedirect.com/science/chapter/handbook/abs/pii/S0927050705801720},
    doi = {10.1016/S0927-0507(05)80172-0},
    issn = {0927-0507}
}

@article{Chauhan2018ConvolutionalRecognition,
    title = {{Convolutional Neural Network (CNN) for Image Detection and Recognition}},
    year = {2018},
    journal = {ICSCCC 2018 - 1st International Conference on Secure Cyber Computing and Communications},
    author = {Chauhan, Rahul and Ghanshala, Kamal Kumar and Joshi, R. C.},
    month = {7},
    pages = {278--282},
    publisher = {Institute of Electrical and Electronics Engineers Inc.},
    url = {https://ieeexplore.ieee.org/abstract/document/8703316},
    isbn = {9781538663738},
    doi = {10.1109/ICSCCC.2018.8703316}
}

@article{Li2017DeepOverview,
    title = {{Deep Reinforcement Learning: An Overview}},
    year = {2017},
    author = {Li, Yuxi},
    month = {1},
    url = {https://arxiv.org/pdf/1701.07274},
    arxivId = {1701.07274}
}

@article{Innan2024FinancialNetworks,
    title = {{Financial fraud detection using quantum graph neural networks}},
    year = {2024},
    journal = {Quantum Machine Intelligence},
    author = {Innan, Nouhaila and Sawaika, Abhishek and Dhor, Ashim and Dutta, Siddhant and Thota, Sairupa and Gokal, Husayn and Patel, Nandan and Khan, Muhammad Al-Zafar and Theodonis, Ioannis and Bennai, Mohamed},
    number = {1},
    pages = {7},
    volume = {6},
    url = {https://doi.org/10.1007/s42484-024-00143-6},
    doi = {10.1007/s42484-024-00143-6},
    issn = {2524-4914}
}

@article{McCaskey2018HybridSystems,
    title = {{Hybrid Programming for Near-Term Quantum Computing Systems}},
    year = {2018},
    journal = {2018 IEEE International Conference on Rebooting Computing, ICRC 2018},
    author = {McCaskey, Alexander and Dumitrescu, Eugene and Liakh, Dmitry and Humble, Travis},
    month = {7},
    publisher = {Institute of Electrical and Electronics Engineers Inc.},
    url = {https://ieeexplore.ieee.org/abstract/document/8638598},
    isbn = {9781538691700},
    doi = {10.1109/ICRC.2018.8638598},
    arxivId = {1805.09279}
}

@article{Gronauer2021Multi-agentSurvey,
    title = {{Multi-agent deep reinforcement learning: a survey}},
    year = {2021},
    journal = {Artificial Intelligence Review 2021 55:2},
    author = {Gronauer, Sven and Diepold, Klaus},
    number = {2},
    month = {4},
    pages = {895--943},
    volume = {55},
    publisher = {Springer},
    url = {https://link.springer.com/article/10.1007/s10462-021-09996-w},
    isbn = {0123456789},
    doi = {10.1007/S10462-021-09996-W},
    issn = {1573-7462}
}

@article{Kraus1997NegotiationEnvironments,
    title = {{Negotiation and cooperation in multi-agent environments}},
    year = {1997},
    journal = {Artificial Intelligence},
    author = {Kraus, Sarit},
    number = {1-2},
    month = {7},
    pages = {79--97},
    volume = {94},
    publisher = {Elsevier},
    url = {https://www.sciencedirect.com/science/article/pii/S0004370297000258},
    doi = {10.1016/S0004-3702(97)00025-8},
    issn = {0004-3702}
}

@article{Barberena2024OverviewMeasurements,
    title = {{Overview of projective quantum measurements}},
    year = {2024},
    author = {Barberena, Diego and Friedman, Aaron J.},
    month = {4},
    url = {https://arxiv.org/pdf/2404.05679},
    arxivId = {2404.05679}
}

@article{Sutton1999PolicyApproximation,
    title = {{Policy Gradient Methods for Reinforcement Learning with Function Approximation}},
    year = {1999},
    journal = {Advances in Neural Information Processing Systems},
    author = {Sutton, Richard S. and McAllester, David and Singh, Satinder and Mansour, Yishay},
    volume = {12}
}

@article{Schulman2017ProximalAlgorithms,
    title = {{Proximal Policy Optimization Algorithms}},
    year = {2017},
    author = {Schulman, John and Wolski, Filip and Dhariwal, Prafulla and Radford, Alec and Openai, Oleg Klimov},
    month = {7},
    url = {https://arxiv.org/pdf/1707.06347},
    arxivId = {1707.06347}
}

@article{Watkins1992Q-learning,
    title = {{Q-learning}},
    year = {1992},
    journal = {Machine Learning 1992 8:3},
    author = {Watkins, Christopher J. C. H. and Dayan, Peter},
    number = {3},
    month = {5},
    pages = {279--292},
    volume = {8},
    publisher = {Springer},
    url = {https://link.springer.com/article/10.1007/BF00992698},
    doi = {10.1007/BF00992698},
    issn = {1573-0565}
}

@article{Yu2023QuantumDirections,
    title = {{Quantum Multi-Agent Reinforcement Learning as an Emerging AI Technology: A Survey and Future Directions}},
    year = {2023},
    journal = {ICCA 2023 - 2023 5th International Conference on Computer and Applications, Proceedings},
    author = {Yu, Wenhan and Zhao, Jun},
    publisher = {Institute of Electrical and Electronics Engineers Inc.},
    url = {https://ieeexplore.ieee.org/abstract/document/10401605},
    isbn = {9798350303254},
    doi = {10.1109/ICCA59364.2023.10401605}
}

@article{Gupta2001QuantumNetworks,
    title = {{Quantum Neural Networks}},
    year = {2001},
    journal = {Journal of Computer and System Sciences},
    author = {Gupta, Sanjay and Zia, R. K.P.},
    number = {3},
    month = {11},
    pages = {355--383},
    volume = {63},
    publisher = {Academic Press},
    url = {https://www.sciencedirect.com/science/article/pii/S0022000001917696},
    doi = {10.1006/JCSS.2001.1769},
    issn = {0022-0000}
}

@article{Dong2008QuantumLearning,
    title = {{Quantum reinforcement learning}},
    year = {2008},
    journal = {IEEE Transactions on Systems, Man, and Cybernetics, Part B: Cybernetics},
    author = {Dong, Daoyi and Chen, Chunlin and Li, Hanxiong and Tarn, Txyh Jong},
    number = {5},
    pages = {1207--1220},
    volume = {38},
    url = {https://ieeexplore.ieee.org/abstract/document/4579244},
    isbn = {9780443302596},
    doi = {10.1109/TSMCB.2008.925743},
    issn = {10834419},
    pmid = {18784007},
    arxivId = {0810.3828}
}

@article{Chen2026QuantumApplications,
    title = {{Quantum reinforcement learning: Concepts and applications}},
    year = {2026},
    journal = {Quantum Computational AI},
    author = {Chen, Samuel Yen-Chi},
    month = {1},
    pages = {3--23},
    publisher = {Morgan Kaufmann},
    url = {https://www.sciencedirect.com/science/chapter/edited-volume/abs/pii/B9780443302596000098},
    doi = {10.1016/B978-0-44-330259-6.00009-8}
}

@article{Chen2025Quantum-Train-BasedLearning,
    title = {{Quantum-Train-Based Distributed Multi-Agent Reinforcement Learning}},
    year = {2025},
    journal = {2025 IEEE Symposium for Multidisciplinary Computational Intelligence Incubators, MCII Companion 2025},
    author = {Chen, Kuan Cheng and Chen, Samuel Yen Chi and Liu, Chen Yu and Leung, Kin K.},
    publisher = {Institute of Electrical and Electronics Engineers Inc.},
    url = {https://ieeexplore.ieee.org/abstract/document/11007471},
    isbn = {9798331519667},
    doi = {10.1109/MCIICOMPANION65207.2025.11007471}
}

@article{Schuld2018SupervisedComputers,
    title = {{Supervised Learning with Quantum Computers}},
    year = {2018},
    author = {Schuld, Maria and Petruccione, Francesco},
    series = {Quantum Science and Technology},
    publisher = {Springer International Publishing},
    url = {https://link.springer.com/10.1007/978-3-319-96424-9},
    address = {Cham},
    isbn = {978-3-319-96423-2},
    doi = {10.1007/978-3-319-96424-9}
}

@article{Cerezo2021VariationalAlgorithms,
    title = {{Variational quantum algorithms}},
    year = {2021},
    journal = {Nature Reviews Physics 2021 3:9},
    author = {Cerezo, M. and Arrasmith, Andrew and Babbush, Ryan and Benjamin, Simon C. and Endo, Suguru and Fujii, Keisuke and McClean, Jarrod R. and Mitarai, Kosuke and Yuan, Xiao and Cincio, Lukasz and Coles, Patrick J.},
    number = {9},
    month = {8},
    pages = {625--644},
    volume = {3},
    publisher = {Nature Publishing Group},
    url = {https://www.nature.com/articles/s42254-021-00348-9},
    doi = {10.1038/s42254-021-00348-9},
    issn = {2522-5820},
    arxivId = {2012.09265}
}

@article{Chen2020VariationalLearning,
    title = {{Variational Quantum Circuits for Deep Reinforcement Learning}},
    year = {2020},
    journal = {IEEE Access},
    author = {Chen, Samuel Yen Chi and Yang, Chao Han Huck and Qi, Jun and Chen, Pin Yu and Ma, Xiaoli and Goan, Hsi Sheng},
    pages = {141007--141024},
    volume = {8},
    publisher = {Institute of Electrical and Electronics Engineers Inc.},
    url = {https://ieeexplore.ieee.org/abstract/document/9144562},
    doi = {10.1109/ACCESS.2020.3010470},
    issn = {21693536},
    arxivId = {1907.00397}
}

\end{document}